\documentclass[10pt,twocolumn,letterpaper]{article}

\usepackage{iccv}
\usepackage{verbatim}
\usepackage{booktabs}
\usepackage{bbm}
\usepackage{tabularx}
\newcolumntype{L}[1]{>{\raggedright\arraybackslash}p{#1}}
\usepackage{color}
\usepackage[accsupp]{axessibility}

\definecolor{iccvblue}{rgb}{0.21,0.49,0.74}
\usepackage[pagebackref,breaklinks,colorlinks,allcolors=iccvblue]{hyperref}

\usepackage{fancyhdr}
\fancypagestyle{ieeefooter}{
  \fancyhf{}

  \fancyfoot[C]{\footnotesize \textcopyright\ 2025 IEEE.  Personal use of this material is permitted.  Permission from IEEE must be obtained for all other uses, in any current or future media, including reprinting/republishing this material for advertising or promotional purposes, creating new collective works, for resale or redistribution to servers or lists, or reuse of any copyrighted component of this work in other works.}
}

\title{Building a General SimCLR Self-Supervised Foundation Model Across Neurological Diseases to Advance 3D Brain MRI Diagnoses}

\author{
Emily Kaczmarek \quad
Justin Szeto \quad
Brennan Nichyporuk \quad
Tal Arbel \\
McGill University, Canada \quad
Mila - Quebec Artificial Intelligence Institute, Canada\\
{\tt\small \{emily.kaczmarek, justin.szeto\}@mail.mcgill.ca} \quad
{\tt\small nichypob@mila.quebec} \quad
{\tt\small tal.arbel@mcgill.ca}
}

\begin{document}
\maketitle

\begin{abstract}
3D structural Magnetic Resonance Imaging (MRI) brain scans are commonly acquired in clinical settings to monitor a wide range of neurological conditions, including neurodegenerative disorders and stroke. While deep learning models have shown promising results analyzing 3D MRI across a number of brain imaging tasks, most are highly tailored for specific tasks with limited labeled data, and are not able to generalize across tasks and/or populations. The development of self-supervised learning (SSL) has enabled the creation of large medical foundation models that leverage diverse, unlabeled datasets ranging from healthy to diseased data, showing significant success in 2D medical imaging applications. However, even the very few foundation models for 3D brain MRI that have been developed remain limited in resolution, scope, or accessibility. In this work, we present a general, high-resolution SimCLR-based SSL foundation model for 3D brain structural MRI, pre-trained on 18,759 patients (44,958 scans) from 11 publicly available datasets spanning diverse neurological diseases. We compare our model to Masked Autoencoders (MAE), as well as two supervised baselines, on four diverse downstream prediction tasks in both in-distribution and out-of-distribution settings. Our fine-tuned SimCLR model outperforms all other models across all tasks. Notably, our model still achieves superior performance when fine-tuned using only 20\% of labeled training samples for predicting Alzheimer's disease. We use publicly available code and data, and release our trained model at  \href{https://github.com/emilykaczmarek/3D-Neuro-SimCLR}{https://github.com/emilykaczmarek/3D-Neuro-SimCLR}, contributing a broadly applicable and accessible foundation model for clinical brain MRI analysis.

\end{abstract}
\thispagestyle{ieeefooter}

\section{Introduction}
\label{sec:intro}

Brain disorders, including neurodegenerative diseases and stroke, are often diagnosed and monitored over time using non-invasive imaging techniques such as 3D structural Magnetic Resonance Imaging (MRI). These images capture detailed anatomical properties of the brain, providing crucial information about features such as the size and shape of brain regions (e.g., ventricle size), as well as pathological features such as lesions and tumours. In recent years, there has been significant success using deep learning to analyze brain structural MRI for a wide variety of individual tasks~\cite{cole, korolev, havaei, chartsias}. However, these architectures are often incredibly tailored to these specific tasks due to the challenges of adapting typical deep learning architectures to limited labeled data. This is a common problem in medical imaging, resulting from (1) time-consuming, expensive labeling processes which must be performed by trained specialists, (2) rare diseases, or (3) small-scale data collection (e.g., clinical trials). These small, customized models trained on specific labels often lack generalizability to other clinical applications, and may even fail on the same task when encountering distribution shifts (e.g., demographic or geographic differences). Some ``larger-scale" models trained with thousands of labeled images have shown promising generalizability, but remain fundamentally limited by the availability of labeled data~\cite{mome2024, wood2024}.

\begin{figure}[h]
    \centering
    \includegraphics[width=0.475\textwidth]{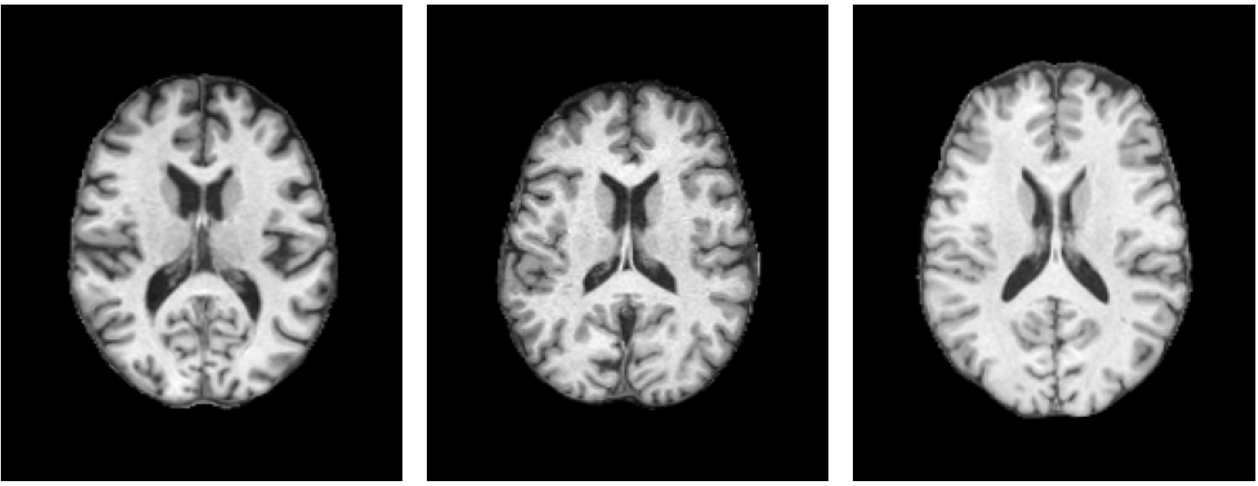}
    \caption{Brain MRI slices of patients with Alzheimer's disease~\cite{ADNI} (left), Parkinson's disease~\cite{PPMI} (centre) and a healthy control~\cite{IXI} (right), showing the similarity of brain images despite different diagnoses.}
    \label{fig:brains}
\end{figure}

In brain imaging, 3D structural MRI scans from different populations share a high degree of visual and anatomical similarity. For example, Figure~\ref{fig:brains} shows brain images of patients diagnosed with Alzheimer's disease, Parkinson's disease, and a healthy control. Rather than being constrained by the limited availability of labeled samples, there exists an unmet opportunity to aggregate diverse datasets spanning both healthy and diseased populations to train large-scale models that learn general representations. Such a model would have a deep understanding of the 3D brain and its components, accounting for wide variability across the population and generalizing well across datasets, disorders, and diagnostic tasks, particularly in situations with limited labeled data available. 

In the field of machine learning, foundation models have gained popularity due to their task-agnostic nature, which allows them to be applied across many diverse tasks. These models are typically trained using Self-Supervised Learning (SSL), where representations are learned by formulating tasks based on the input data alone (rather than external annotations), and are thus capable of leveraging large unlabeled datasets. For example, contrastive-based self-supervised algorithms such as SimCLR \cite{simclr} and MoCo \cite{moco} use 2D image augmentations to create multiple views of the same natural image, and push these representations together to identify high-level semantic features that are invariant to transformations. Other self-supervised models, such as Masked Autoencoders (MAEs) \cite{mae} and the Image-based Joint-Embedding Predictive Architecture (I-JEPA) \cite{jepa} focus on learning semantically rich representations through mask prediction tasks. 

In the field of medical imaging, SSL foundation models are in their infancy. Several 3D brain MRI foundation models have been trained using self-supervision, focusing on downstream objectives such as brain lesion segmentation \cite{brainseg}, brain tumour detection \cite{lamim}, or a small variety of tasks including brain age regression, anatomical segmentation, and lesion detection \cite{karimi}, or schizophrenia, bipolar disorder, and Alzheimer's classification \cite{dufumier}.  However, current existing 3D brain MRI foundation models are severely limited. First \textbf{(1)}, while a small number of foundation models have recently been developed exclusively for brain structural MRI analysis, most limit their training data to one to two datasets and/or pathologies \cite{brainseg, karimi, dufumier}.
While these models may exhibit strong in-distribution performance tailored to specific pathologies, ideally, there should also be foundation models trained on large-scale diverse data to improve adaptability across many clinical brain imaging tasks. Second \textbf{(2)}, many models that are currently released are low resolution (or significantly cropped) due to the computational demands of processing high-dimensional 3D medical images \cite{lamim, dufumier}. Most 3D brain MRI scans are acquired at a resolution of $1 \times 1 \times 1~\mathrm{mm}^3$, which enables the identification of small, clinically relevant features. For example, lesions can be as small as three voxels, which may be blurred, lost, or undetectable at lower resolutions. This highlights the importance of training large-scale, high-resolution foundation models for 3D medical imaging applications. Third \textbf{(3)}, while some studies have developed generalizable models trained using self-supervised learning, the models may remain unreleased \cite{karimi} or use complicated preprocessing pipelines that are not easily reproducible \cite{brainseg}. As a result, these models fail to meet the requirements for a true foundation model that is generalizable and easy to use for numerous downstream tasks.

In this work, we address these three limitations of current foundation models by presenting a high-resolution, generalizable foundation model for 3D brain structural MRI analysis across neurological diseases, using fast and publicly available preprocessing tools to enable quick adaptation to new datasets:
\begin{itemize}
\item We develop a high-resolution, diverse 3D brain structural MRI foundation model based on the SimCLR architecture, trained on the aggregation of 11 different datasets with 18,759 patients (totaling 44,958 scans) spanning numerous neurological conditions including Alzheimer's disease, Parkinson's disease, and stroke. We also implement and compare against the state-of-the-art MAE architecture.
\item We evaluate our foundation model against state-of-the-art supervised models, tested across four diverse downstream tasks (both in-distribution and out-of-distribution) for healthy and sick patients.
\item We perform experiments to determine the minimal amount of data required to reach optimal performance, assessing adaptability to the extremely small datasets common in medical imaging.
\item We release our code and trained SimCLR-based foundation model for public use on GitHub.
\end{itemize}

We first define our contrastive, Convolutional Neural Network (CNN)-based SimCLR approach and compare against a masking-based Vision Transformer (ViT) to determine the optimal configuration for 3D brain MRI analysis. We find that the CNN-based SimCLR model performs exceptionally well across all tasks;  our results show that the fine-tuned SimCLR 3D brain structural MRI foundation model outperforms all supervised and self-supervised models for both in-distribution and out-of-distribution tasks. Importantly, we discover that the SimCLR model achieves comparable performance to supervised methods when fine-tuning on only 20\% of the data. 

\section{Related Work}
\label{sec:related_work}

There currently exist numerous large-scale foundation models both in natural imaging~\cite{caron2021emerging,mae,jepa} and medical imaging~\cite{perezfoundation,azizi,MedSAM}. However, these models are typically developed for 2D images, which are far more prevalent and require less compute for analysis, or they process 3D images as individual slices and therefore lose their spatial consistency. Here we focus our review on large-scale models trained on 3D brain MRI scans. Specifically, we identify two main areas of research: large-scale supervised models developed for specific medical tasks (including those referred to as foundation models), and self-supervised foundation models designed to be more generally applicable across tasks. In particular, we discuss the size and diversity of the training/testing data, as well as the image resolution.

\subsection{Large-Scale Supervised Models}
Supervised learning is the main approach used to analyze 3D brain structural MRI. Recently, a number of ``foundation models" have been developed by training on specific tasks using supervised objectives. These are described as foundation models as they leverage larger-scale 3D brain MRI datasets, either from a single source or by merging several (small) datasets, despite being tailored to narrow applications. Wood et al. trained five different models for individual MRI sequences (i.e., unique MRI scans of the same patient that highlight different brain features) for brain age estimation, using between 7,000-18,000 healthy scans per model, and tested on out-of-distribution age prediction of Alzheimer's patients \cite{wood2024}. This study used images with lower resolution ($1.4~\mathrm{mm}^3$). \textit{MoME} was developed for brain lesion segmentation and consists of five individual nnU-Net models (each trained on a single MRI sequence) with a gating network that combines predictions across desired sequences~\cite{mome2024}. Training and testing were performed on subsets of a dataset containing 6,585 patients with eight different brain lesion pathologies.

While large-scale supervised approaches exist, their reliance on labels limits their scalability. Deep learning models benefit from larger amounts of data~\cite{chen2017}, but label availability constrains the number of 3D brain MRI scans that can be used for training. Even commonly available labels, such as age, can introduce ambiguity; for example, a younger patient with Alzheimer’s disease may exhibit brain features similar to those of a healthy older adult. This limits the interpretability of age and effectively reduces the size of the usable training set.

\subsection{Brain MRI Foundation Models}
In representation learning, foundation models are typically trained using self-supervised learning, which allows the aggregation of large, diverse, unlabeled datasets.
Recently (at the time of this paper), four 3D brain structural MRI foundation models have been published. These have been pre-trained entirely with self-supervised learning: (1)  {\it BrainSegFounder} trained a Vision Transformer encoder using 41,400 healthy patients with both T1 and T2 MRI sequences \cite{brainseg}. The model was then fine-tuned, via a decoder, for the tasks of brain tumour and stroke lesion segmentation. (2) {\it LaMIM} used a Masked Autoencoder (MAE) on a private dataset consisting of 57,621 patients with four different MRI sequences, and evaluated on three brain tumour classification tasks \cite{lamim}. Despite training with self-supervised learning for pre-training and therefore being applicable across tasks, both of these models focused entirely on segmentation or detection tasks for a maximum of two different pathologies. (3) Karimi trained a network with a novel self-supervised learning strategy using 11,000 structural MRI and 12,000 diffusion MRI scans \cite{karimi} aggregated from 10 public datasets (primarily healthy patients). This model was evaluated on more tasks than the previous two studies, including brain age regression, anatomical segmentation, and lesion detection. However, to the best of our knowledge, this model is not publicly available, which reduces its usability as a general purpose brain structural foundation model. 
(4) Although not self-titled as a foundation model, Dufumier et al. trained a self-supervised SimCLR model on $10^4$ healthy brain samples, and evaluated it on schizophrenia, bipolar disorder, and Alzheimer's prediction tasks \cite{dufumier}. Their model was tested on a range of tasks and released to the public, but relied on images at a lower resolution ($1.5~\mathrm{mm}^3$).  
In addition to the weaknesses outlined \textit{per study} above, we note that while each work provides details on preprocessing, many are missing precise details for exact reproducibility. This may cause reduced performance when attempting to apply these models more broadly, and highlights the need for using standardized preprocessing pipelines in training foundation models. Furthermore, the majority of these models have only been trained on one to two populations and/or pathologies (e.g., all healthy patients, all tumour patients), and have not taken advantage of the ability to combine datasets for large-scale training.

Overall, there is a lack of publicly available, easy-to-use, diverse, high-resolution foundation models that are generalizable to different contexts of 3D brain MRI analysis. 

\section{Methodology}
\label{sec:methodology}
Our overall goal is to develop a  foundation model that is able to learn generalizable 3D brain MRI representations for diverse downstream tasks. Given the subtle anatomical changes across subjects and/or disease states in brain MRI, we adapt a CNN-based SimCLR architecture (selected for its strong inductive biases, given our smaller dataset relative to natural imaging) to handle 3D inputs.  Additionally, we also adapt a Masked Autoencoder (MAE) strategy as a complementary self-supervised baseline to evaluate alternative pre-training objectives.

\begin{figure}[h]
    \centering
    \includegraphics[width=0.475\textwidth]{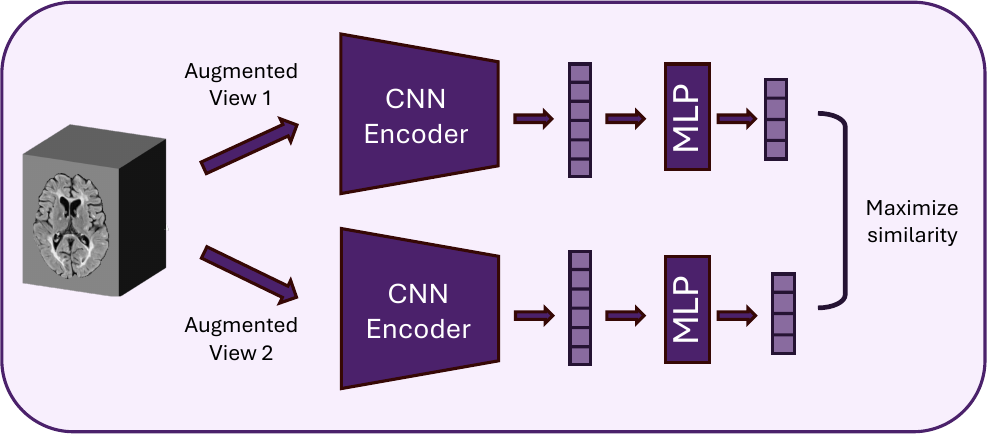}
    \caption{SimCLR Architecture. Two different views of a single 3D MRI scan are created through random augmentations. These are used  as input to the same encoder to create separate representations. The representations are then projected into a lower dimension, where similarity between the two projections of the same image is maximized (and minimized between all projections from other images). Inspired by \cite{simclr}.}
    \label{fig:simclr}
\end{figure}

\subsection{SimCLR for 3D Brain MRI}
We adapt the SimCLR framework for self-supervised representation learning with 3D brain MRI (Figure \ref{fig:simclr}). SimCLR is a contrastive learning-based algorithm that aims to increase similarity between representations from different augmented views of the same image, while decreasing similarity with views of other images. Given a 3D image $x$, two different augmentations are applied to the image, $x_i$ = $t_i(x)$ and $x_j$ = $t_j(x)$. These augmented images are then used as input to a convolutional encoder to produce latent representations, $h_i = f(x_i)$ and $h_j = f(x_j)$. Lastly, these representations are passed through a projection head to produce the final outputs, $z_i = g(h_i)$ and $z_j = g(h_j)$. To learn high-level semantic representations that are invariant to the selected data augmentations, the similarity between the final projections is maximized through the following loss function (normalized temperature-scaled cross-entropy, NT-Xent \cite{simclr}): 
\begin{equation}
\mathcal{L}_{i,j} = -\log \frac{\exp(\mathrm{sim}(z_i, z_j) / \tau)}{\sum_{k=1}^{2N} \mathbbm{1}_{[k \ne i]} \exp(\mathrm{sim}(z_i, z_k) / \tau)}
\label{eq:simclr-loss}
\end{equation}
where $\text{sim}(z_i, z_j) = \frac{z_i^\top z_j}{\|z_i\|\|z_j\|}$ is the cosine similarity between projections and $\tau$ is a temperature parameter. The total loss is computed by averaging $\mathcal{L}_{i,j}$ and $\mathcal{L}_{j,i}$ for each positive pair (i.e., views of the same image), and then averaging these per-pair losses over the batch. Extending SimCLR to 3D brain images involves replacing the 2D CNN with a 3D CNN, and adapting the original SimCLR augmentations for 3D volumes (discussed further in Section \ref{sec:models}). After pre-training, image representations are extracted from the encoder $f(\cdot)$ for downstream classification or regression tasks.

\begin{figure}[h]
    \centering
    \includegraphics[width=0.475\textwidth]{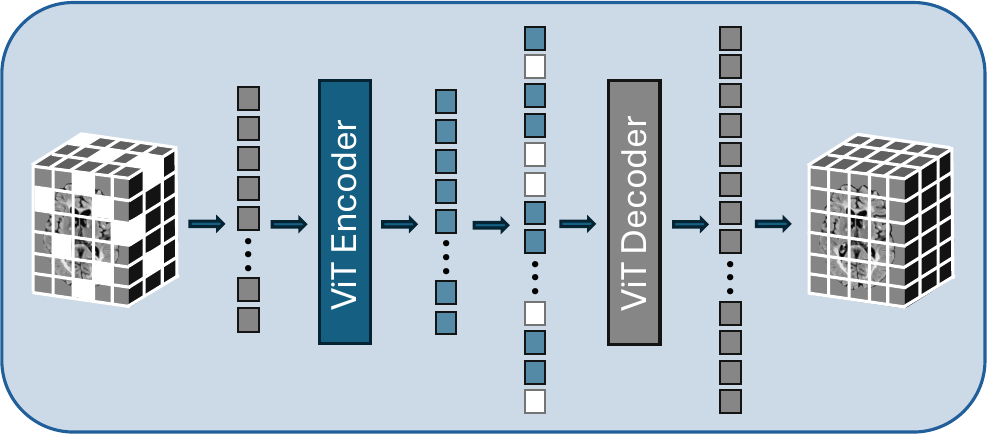}
    \caption{MAE Architecture. First, 3D patches from MRI volumes are randomly masked. The unmasked patches are then projected into linear tokens, which are used as input to the ViT encoder. The mask tokens are then concatenated with the unmasked representations, and subsequently predicted by the ViT decoder. Inspired by~\cite{mae}.}
    \label{fig:mae}
\end{figure}

\subsection{MAE for 3D Brain MRI}
We implement a 3D Masked Autoencoder (MAE) to learn image representations by reconstructing missing patches from 3D brain MRI scans (Figure \ref{fig:mae}). Briefly, input images are first divided into non-overlapping patches, and a certain percentage of the patches are masked out. 
In MAE, the masked patches are removed entirely from the encoder input, resulting in a lightweight encoder that processes only the unmasked patches. These unmasked patches are linearly projected into 1D embeddings and fed into the Vision Transformer (ViT) encoder. For reconstruction, mask tokens (which indicate the locations to reconstruct) are concatenated with the encoded unmasked patch representations. This combined sequence is then passed to a ViT decoder, which predicts the masked patches in voxel space. The model is trained using Mean Squared Error (MSE) on the masked patch reconstructions:
\begin{equation}
\mathcal{L}_{\text{MAE}} = \frac{1}{|M|} \sum_{i \in M} \| \hat{x}_i - x_i \|_2^2
\label{eq:mae-loss}
\end{equation}
where $M$ is the set of masked patch indices, $x_i$ is the original patch, and $\hat{x}_i$ is the predicted patch. After pre-training, the ViT encoder is used for downstream classification or regression tasks.

\section{Experiments and Implementation Details}
\label{sec:experiments}
To develop our foundation model, we aggregate data from 11 different sources for self-supervised pre-training. We then evaluate model performance using two approaches: in-distribution and out-of-distribution. Specifically, for in-distribution, we divide one of our datasets into training, validation, and test splits \textit{prior} to pre-training with SSL. The held-out validation and test data are used to assess model performance within the same data distribution (i.e., acquisition source). Next, for out-of-distribution evaluation, we completely hold out two datasets during training to evaluate the model's ability to adapt to data acquired from different sources. All preprocessing steps are identical across datasets, and the pre-trained SimCLR and MAE models are compared with supervised models trained from scratch.

\subsection{Datasets}
\textbf{Training Datasets}
We implement all of our models with 3D MRI brain scans aggregated from 11 datasets, shown in detail in Table \ref{tab:datasets}. To account for different MRI sequences  (i.e., brain scans using different MRI settings to highlight unique brain features) being available in each dataset, we select the T1 sequence from all datasets, which is the most common. In total, we pre-train the SSL models on 18,759 patients and 44,958 scans (considering that each patient may have more than one scan). The conditions listed in Table \ref{tab:datasets} are the primary diagnostic categories considered in each study. However, some datasets contain patients with additional underrepresented diseases and/or disorders, which may be caused by randomized recruiting, or enrolling individuals with similar symptoms. For example, the HABS-HD \cite{HABS_HD} dataset focuses on healthy individuals and patients with mild to full dementia, while also containing mental health disorders such as depression and anxiety, and individuals with a history of traumatic brain injury. Similarly, while PPMI \cite{PPMI} primarily contains patients diagnosed with Parkinson's disease, there are also a very small number patients who have been diagnosed with more rare neurological disorders such as corticobasal syndrome and spinocerebellar ataxia. These datasets also contain diverse demographics; for example, HABS-HD \cite{HABS_HD} was specifically designed to ensure ethnic diversity among participants.

\begin{table}[t]
\centering
\begin{tabular}{lccl}
\toprule
\textbf{Dataset} & \textbf{\# Patients} & \textbf{\# T1 Scans} & \textbf{Conditions}  \\
\midrule
ADNI \cite{ADNI}    & 3,028  & 21,778  & H, MCI, AD         \\
AOMIC \cite{AOMIC}& 1,368    & 1,368  & H             \\
CoRR \cite{CoRR} & 1,416    & 1,416    & H         \\
DLBS \cite{DLBS} & 315  & 315  & H         \\
GSP \cite{GSP} & 1,569    & 1,569   & H          \\
HABS-HD \cite{HABS_HD} & 4,231  & 6,429  & H, MCI, AD                 \\
MCSA \cite{MCSA} & 1,802  & 3,090  & H, MCI, AD                 \\
NIFD \cite{NIFD}& 346  & 1,815  & H, FTD             \\
PPMI \cite{PPMI}& 3,124    & 5,618    & H, PD           \\
SALD \cite{SALD} & 458 & 458 &H \\
SOOP \cite{SOOP}  & 1,102    & 1,102   & H, S         \\
\midrule
\textbf{Train total} & \textbf{18,759} & \textbf{44,958} & -- \\
\midrule
\multicolumn{4}{l}{\textit{Held-out test datasets}} \\
\midrule
AIBL \cite{AIBL}& 704    & NA    & H, MCI, AD         \\
IXI \cite{IXI} & 582    & NA    & H              \\
SOOP \cite{SOOP}& 602    & NA    & H, S          \\
\bottomrule
\end{tabular}

\caption{Summary of the 11 publicly available datasets used for pre-training. SOOP was split between pre-training and evaluation, while AIBL and IXI were used exclusively for held-out downstream testing. H, Healthy Individuals; MCI, Mild Cognitive Impairment; AD, Alzheimer's Disease; FTD, Frontotemporal Dementia; PD, Parkinson's Disease; S, Stroke; NA, Not Applicable.}
\label{tab:datasets}
\end{table}

\noindent\textbf{Test Datasets}
We first evaluate the model on one in-distribution task using the SOOP dataset.

\noindent\textit{Stroke Outcome Optimization Project (SOOP)}: The SOOP dataset contains 1,704 patients who have experienced a stroke. For this task, we perform regression on the National Institutes of Health (NIH) Stroke Scale, which ranges from 0 to 42, where a score of 0 indicates no symptoms, 1-4 indicates a minor stroke, 5-15 indicates a moderate stroke, 16-20 indicates a moderate to severe stroke, and 20-42 indicates a severe stroke. We use this dataset for our in-distribution evaluation with a 60/20/20 split for training/validation/test, and use the training data for both SSL pre-training and downstream fine-tuning. All validation and test data are completely held-out during pre-training and only used for downstream evaluation.

\noindent\textit{Australian Imaging, Biomarkers and Lifestyle (AIBL):} The AIBL dataset consists of longitudinal images from healthy patients, as well as those with Mild Cognitive Impairment (MCI) and Alzheimer's Disease (AD). We reserve this dataset for out-of-distribution evaluation to determine how well the model generalizes to unseen data, particularly in the context of disease classification. We perform binary classification of healthy vs Alzheimer's patients. We balance the training set with a total of 106 patients, and keep the existing data imbalance in the validation and test sets, yielding 215 healthy / 15 Alzheimer's patients in validation and 216 healthy / 15 Alzheimer's patients in the test set.

\noindent\textit{Information eXtraction from Images (IXI):} The IXI dataset contains 582 healthy control subjects. This dataset is also held-out for out-of-distribution evaluation to assess how well the pre-trained model captures healthy brain characteristics in an independent cohort. The data are divided into 60/20/20 training/validation/test splits.

\noindent\textbf{Preprocessing}
For preprocessing, we use the publicly available TurboPrep package on GitHub\footnote{\url{https://github.com/LemuelPuglisi/turboprep}}. With TurboPrep, we perform standard MRI preprocessing which includes: N4 bias-field correction to reduce scanner-related intensity inhomogeneities \cite{n4};  skull stripping to remove non-brain voxels from the image \cite{skull}; linear registration to align all scans to a common template with standardized brain orientation and size \cite{registration}; brain region segmentation to outline and identify specific anatomical structures \cite{segmentation}; and intensity normalization to standardize voxel intensities across subjects \cite{normalization}. Skull stripping and brain region segmentation are performed in TurboPrep using trained deep learning algorithms for improved efficiency. On average, the entire preprocessing pipeline takes approximately one minute per scan, which is relatively fast and completely reproducible. All volumes are resampled to $1 \times 1 \times 1~\mathrm{mm}^3$ (for high-resolution), corresponding to a voxel size of 193x229x193. We apply a permutation so the first dimension corresponds to image depth, and center crop of 150x192x192 to remove background-only slices and peripheral voxels that do not contain brain tissue. Lastly, we perform Z-score normalization on each individual sample.

\begin{table*}[t]
\centering
\begin{tabular}{l cc cc cc cc cc}
\toprule

& \multicolumn{1}{c}{\textbf{In-Distribution}} 
& \multicolumn{3}{c}{\textbf{Out-of-Distribution}} \\
\cmidrule(lr){2-2} \cmidrule(lr){3-5}

\textbf{Model/Task}  &\textbf{Stroke Scale Regression}& \textbf{Alzheimer’s Classification} & \textbf{Sex Classification} & \textbf{Age Regression}  \\
Metric & Mean Absolute Error $\downarrow$& AUC $\uparrow$ & AUC $\uparrow$ & Mean Absolute Error $\downarrow$ \\
\midrule

\textit{Supervised} \\
ResNet-18   &   5.47  $\pm$ 0.09  &  0.869 $\pm$ 0.030     &  0.988 $\pm$ 0.005     &  4.85 $\pm$ 0.13          \\

ViT-T & 6.02  $\pm$ 0.09  & 0.844 $\pm$ 0.054  &  0.876 $\pm$  0.014 &	7.53 $\pm$ 0.25        \\

\textit{Self-Supervised} \\
MAE-LP   & 6.12   $\pm$   0.02  &    0.539 $\pm$  0.053       &  0.839   $\pm$ 0.026 & 15.28  $\pm$  0.18               \\

MAE-FT& 6.15 $\pm$ 0.15  &	0.798 $\pm$	0.042  & 0.864 $\pm$ 0.025 &	8.53 $\pm$ 0.28       \\

SimCLR-LP  & 5.87 $\pm$      0.07  & 0.904   $\pm$  0.008  &   0.896  $\pm$  0.006  &  5.90  $\pm$ 0.16          \\        
SimCLR-FT & \textbf{5.37 $\pm$ 0.24} & \textbf{0.929 $\pm$ 0.028}  & \textbf{0.991 $\pm$ 0.004} &   \textbf{4.35 $\pm$ 0.22}             \\  
\bottomrule
\end{tabular}
\caption{Performance (AUC / Mean Absolute Error) of all models across four downstream tasks: stroke scale regression (SOOP dataset), classification of Alzheimer’s disease against healthy controls (AIBL dataset), sex classification (IXI dataset), and age regression (IXI dataset). $\uparrow$ indicates higher values are better, while $\downarrow$ indicates lower values are better. The mean absolute error for the stroke scale regression task is reported in terms of the NIH stroke scale, and reported in years for the age regression task. We demonstrate performance across in-distribution tasks (stroke scale regression) and out-of-distribution tasks. Importantly, the fine-tuned SimCLR foundation model (SimCLR-FT) outperforms all other models on every task, highlighting the importance of developing large-scale models for 3D brain MRI prediction tasks. Results are reported as the mean $\pm$ standard deviation across five independent runs. LP, Linear Probe; FT, Fine-tuned.}
\label{tab:main_results}
\end{table*}

\subsection{Model Architectures}
\label{sec:models}

We compare two baselines against our pre-trained self-supervised models. All models are trained on the same preprocessed (downstream) datasets and evaluated on identical splits for consistency. For pre-training on SimCLR and MAE, each patient is seen once per epoch, where the input scan is sampled from their available scans. This ensures that patients with more scans do not overrepresent the training data. During evaluation, the SimCLR and MAE models are evaluated with both linear probing and fine-tuning for 100 epochs, while supervised models are trained for 300 epochs.

\noindent\textbf{Baselines:} We compare the pre-trained SSL models against state-of-the-art supervised learning models. Specifically, we choose to compare against the same architectures used as the backbones of our self-supervised models, allowing the comparison between CNNs and ViTs trained from scratch. We therefore compare against a ResNet-18 \cite{resnet}, implemented from the same repository as our SimCLR model, and a ViT-Tiny \cite{vit} baseline from the same repository as our MAE model. All models are trained using standard data augmentations adapted for 3D grayscale images, implemented with the Medical Open Network for Artificial Intelligence (MONAI \cite{monai}), which provides public data augmentations, code, and models for medical imaging applications: \texttt{RandSpatialCropd} with random (minimum) size  $(90 \times 115 \times 115)$, followed by \texttt{Resized} to $150 \times 192 \times 192$; \texttt{RandFlipd} along the axial axis ($p=0.5$); \texttt{RandAffined} with rotation range $(0.1, 0.1, 0.1)$~rad, scale range 15\%, and translation range 5 voxels ($p=0.7$); \texttt{RandShiftIntensityd} with offset 0.1 ($p=0.5$); and \texttt{RandGaussianNoised} with std$=$0.1 ($p=0.2$).

\noindent\textbf{SimCLR:} We implement SimCLR using the publicly available GitHub repository\footnote{\url{https://github.com/Spijkervet/SimCLR}}. The backbone of our SimCLR model is a ResNet-18, where the final layer of the model is replaced by a projection head of dimension 64. We use a series of strong 3D augmentations implemented with MONAI \cite{monai}: \texttt{RandSpatialCropd} with random center and (minimum) size $(30 \times 40 \times 40)$, followed by \texttt{Resized} to $150 \times 192 \times 192$; \texttt{RandFlipd} along the axial axis ($p=0.5$); \texttt{RandRotated} with a rotation range of 45° ($p=0.5$); \texttt{RandShiftIntensityd} with offset 0.5 ($p=0.8$); and \texttt{RandAdjustContrastd} with gamma range (0.5, 1.5) ($p=0.8$). We train SimCLR on 12 H100 GPUs with an effective batch size of 72 for 150 epochs. For downstream evaluation, no augmentations are used.

\noindent\textbf{MAE:} Our MAE model is achieved using the publicly available GitHub repository\footnote{\url{https://github.com/facebookresearch/mae}}. We choose 3D patch sizes of $25\times16\times16$, resulting in 864 patches in total, and add sinusoidal positional embeddings to each patch representation. Three different ViT sizes are trained to determine the optimal selection for 18,759 3D brain MRIs: ViT-Tiny, ViT-Base, and ViT-Large. We notice that the performance of all three models are similar, and show our performance on ViT-Tiny due to its lower computational requirements and reduced risk of overfitting. We use MONAI \cite{monai} \texttt{RandSpatialCropd} with random center and (minimum) size $(30 \times 40 \times 40)$, followed by \texttt{Resized} to $150 \times 192 \times 192$; \texttt{RandFlipd} along the axial axis ($p=0.5$) augmentations during pre-training, and apply the same augmentations as the supervised baselines during downstream fine-tuning (no augmentations are applied for linear probing). We keep the original hyperparameters of MAE and mask 75\% of tokens. We train the MAE models on 8 H100 GPUs with an effective batch size of 128 for 150 epochs.

\section{Results}
\label{sec:results}

\subsection{Comparison of Supervised and Self-Supervised Models for 3D Brain MRI Tasks}

\textbf{Overall trends in performance}
Table \ref{tab:main_results} shows the performance of the self-supervised models and their supervised comparisons across four downstream tasks. Importantly, the fine-tuned SimCLR model outperforms all other models (supervised ResNet-18, ViT-T, and self-supervised MAE, as well SimCLR-based linear probing) on all tasks. This highlights the importance of developing foundation models for 3D brain MRI: high-dimensional data with limited labeled samples creates a challenging learning setting that can be improved by training large-scale, general-purpose models. Additionally, we demonstrate that a single model is able to be fine-tuned across a range of tasks, avoiding the need to develop separate, highly specialized models for each application. Notably, despite being fine-tuned for only 100 epochs, SimCLR outperforms supervised baselines trained for 300 epochs, demonstrating both effectiveness and computational efficiency.

The two models trained using ViT architectures (ViT-T and MAE) struggle to match the performance of the CNN models (ResNet-18 and SimCLR). This is a well-known problem: ViTs excel when large amounts of data are available. While our dataset of 18,759 patients is substantial for medical imaging, it still remains small compared to natural imaging datasets, and therefore it is difficult for these models to outperform CNNs with strong inductive biases. In addition, ViTs are more sensitive to specific hyperparameters, such as patch size and model depth. Despite these challenges, several studies have achieved state-of-the-art performance using MAEs on other 3D brain MRI tasks such as brain tumor classification and segmentation \cite{lamim, brainseg}. The purpose of this study is to develop a publicly available, high-performing foundation model, which we achieve using SimCLR. We include our MAE results to illustrate the challenges of training ViTs from scratch, and leave further optimization of this architecture for future work. 

\textbf{In-Distribution Performance}
First, we evaluate a single in-distribution task, where 60\% of the SOOP dataset (containing stroke patients) was used for SSL pre-training, while 40\% were held-out for evaluation. While predicting the NIH stroke score is inherently challenging and noisy, the SimCLR fine-tuned model has superior performance over all other models, as shown by column 1 in Table~\ref{tab:main_results}. 

\textbf{Out-of-Distribution Performance}
Next, we evaluate three out-of-distribution tasks to determine if the model has learned generalizable features that extend to previously unseen datasets. The first task is the classification of Alzheimer's patients against healthy controls, shown in column 2 of Table \ref{tab:main_results}. The fine-tuned SimCLR model particularly excels here, achieving a 6\% AUC improvement in classifying Alzheimer's disease over the next-best model. Importantly, this training dataset contains only 106 patients (53 per class), demonstrating the strong performance of foundation models in situations with limited labeled data. 

The IXI dataset consists of healthy individuals, making it ideal for tasks such as sex classification and age regression to demonstrate if the models have a strong understanding of overall brain characteristics. The SimCLR model is able to achieve near-perfect accuracy on sex classification, and outperforms all other models on age regression, indicating that the model has successfully identified important brain features that generalize to out-of-distribution samples  (Table \ref{tab:main_results}, columns 3 and 4).

\begin{figure}[h]
    \centering
    \includegraphics[width=0.45\textwidth]{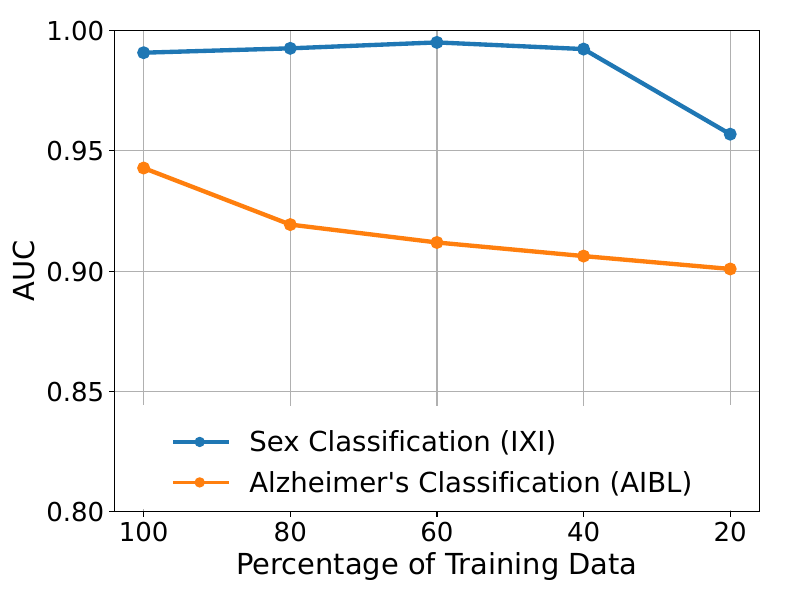}
    \caption{Classification performance (AUC) of two downstream tasks as the percentage of training data decreases when fine-tuning SimCLR. The SimCLR-based foundation model is able to maintain high performance as data decreases, outperforming all other models (trained using all data) on classifying Alzheimer's disease (orange, from the AIBL dataset). Both classification tasks are out-of-distribution.}
    \label{fig:classification}
\end{figure}

\begin{figure}[h]
    \centering
    \includegraphics[width=0.45\textwidth]{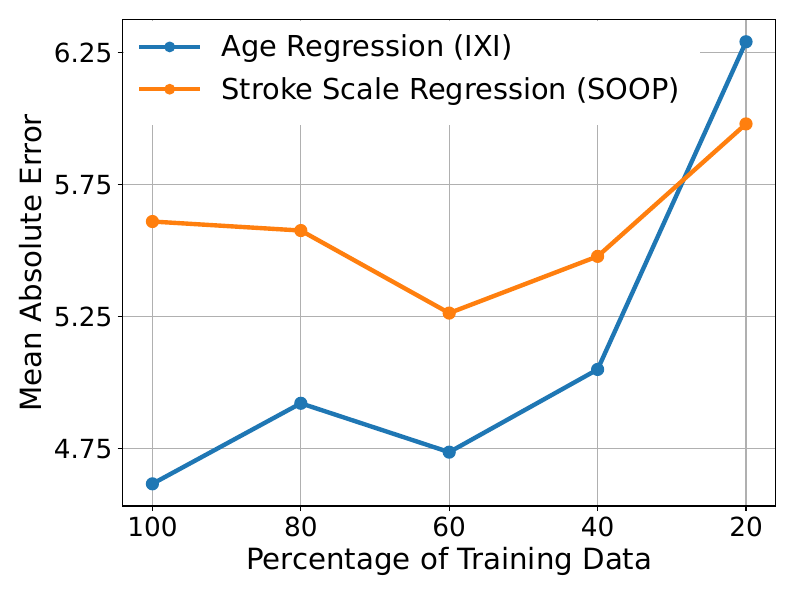}
    \caption{Performance (Mean Absolute Error) of age (blue, out-of-distribution from the IXI dataset) and stroke scale (orange, in-distribution from the SOOP dataset) regression as the percentage of training data decreases when fine-tuning SimCLR, where lower error values are better. The MAE remains low until the training dataset is reduced to 20\% of its original size.}
    \label{fig:regression}
\end{figure}

\subsection{Percentage of Training Data Available}
Many 3D brain MRI datasets have a limited amount of labeled samples, due to factors such as the need for expert and time-consuming annotations, rare diseases, and small-scale clinical trials. We therefore evaluate how much training data is required to fine-tune SimCLR and still achieve optimal performance across all tasks. Figures \ref{fig:classification} and \ref{fig:regression} show the AUC and mean absolute error, respectively, of each of the four tasks when trained on 20\%, 40\%, 60\%, 80\%, and 100\% of the available training data (each model is trained once using the same seed for comparison). Most tasks show minimal decrease in performance until the training set is reduced to 20\% of its original size. For the classification of Alzheimer's against healthy controls, the AUC drops by 4\% over the course of reducing the training data. 
The validation and test sets of this dataset have substantial class imbalances and would likely benefit from greater positive-sample representation in the training set.
However, with only 20\% of data (10 samples per class for Alzheimer's prediction), the model is still able to outperform all supervised models for this task. These results demonstrate the strength of our 3D brain MRI foundation model; even in settings with limited labeled data and challenging disease prediction tasks, it is able to excel.

\section{Conclusion}
\label{sec:conclusion}
In this work, we present a publicly available, high-resolution, generalizable 3D brain MRI SSL foundation model. Importantly, we demonstrate that our foundation model outperforms supervised learning across four different tasks, showing applicability across numerous diseases and usefulness for situations with limited annotated data. We also find that our model continues to outperform supervised learning with only 20\% of labeled training data. Our model and code are released publicly, and use publicly available, fast preprocessing. Overall, with its release, the model will permit generalization to a wide variety of 3D brain imaging tasks, achieving high performance results in real-world clinical contexts even when only extremely limited data is available.

\section*{Acknowledgements}
This work was supported by the Natural Sciences and Engineering Research Council of Canada, Fonds de Recherche du Quebec: Nature et Technologies, the Canadian Institute for Advanced Research (CIFAR) Artificial Intelligence Chairs program, Calcul Quebec, the Digital Research Alliance of Canada, the Vadasz Scholar McGill Engineering Doctoral Award, Mila - Quebec AI Institute, the International Progressive MS Alliance and the MS Society of Canada.

We also acknowledge all datasets used in this work. AIBL: Data used in the preparation of this article was obtained from the Australian Imaging Biomarkers and Lifestyle flagship study of aging (AIBL) funded by the Commonwealth Scientific and Industrial Research Organisation (CSIRO). ADNI: Data collection and sharing for the Alzheimer's Disease Neuroimaging Initiative (ADNI) is funded by the National Institute on Aging (National Institutes of Health Grant U19AG024904). GSP: Data were provided [in part] by the Brain Genomics Superstruct Project of Harvard University and the Massachusetts General Hospital, with support from the Center for Brain Science Neuroinformatics Research Group, the Athinoula A. Martinos Center for Biomedical Imaging, and the Center for Human Genetic Research. HABS-HD: Research reported on this publication was supported by the National Institute on Aging of the National Institutes of Health under Award Numbers R01AG054073, R01AG058533, P41EB015922 and U19AG078109. MCSA: The data contained in this analysis were obtained under research grant from the National Institutes of Health to the Mayo Clinic Study of Aging (U01 AG006786, Ronald Petersen, PI). NIFD: Data collection and sharing for this project was funded by the Frontotemporal Lobar Degeneration Neuroimaging Initiative (National Institutes of Health Grant R01 AG032306). PPMI: Data used in the preparation of this article was obtained on [2025-05-05] from the Parkinson's Progression Markers Initiative (PPMI) database. PPMI – a public-private partnership – is funded by the Michael J. Fox Foundation for Parkinson's Research, and funding partners. We additionally thank the contributors and data curators of the Amsterdam Open MRI Collection (AOMIC), Consortium for Reliability and Reproducibility (CoRR), Dallas Lifespan Brain Study (DLBS), Information eXtraction from Images (IXI), Southwest University Adult Lifespan Dataset (SALD), and Stroke Outcome Optimization Projection (SOOP), whose efforts made this research possible.
{
    \small
    \bibliographystyle{ieeenat_fullname}
    \bibliography{main}
}

\end{document}